\documentclass[11pt]{article}
\usepackage{setspace}
\usepackage{amsfonts}
\usepackage{amssymb}         
\usepackage{amsmath}    
\usepackage{multirow}
\usepackage{graphicx}
\usepackage{algorithm}
\usepackage{algorithmic}

\newtheorem{thm}{Theorem}

\newtheorem{remark}[thm]{Remark}

\newtheorem{definition}[thm]{Definition}

\begin{document}

\title{On the Evaluation Criterions for the Active Learning Processes}
\author{\textsc{Vladimir Nikulin\thanks{Email: vnikulin.uq@gmail.com}}    \\ 
Department of Mathematics, University of Queensland \\
Brisbane, Australia}

\date{}
\maketitle
\thispagestyle{empty}

\begin{abstract}
In many data mining applications collection of sufficiently large datasets 
is the most time consuming and expensive.
On the other hand, industrial methods of data collection create huge databases, and 
make difficult direct applications of the advanced machine learning algorithms.
To address the above problems, we consider active learning (AL), 
which may be very efficient either for the experimental design or for the data filtering.
In this paper we demonstrate using the online evaluation opportunity 
provided by the AL Challenge that quite competitive results may be produced 
using a small percentage of the available data. 
Also, we present several alternative  criteria, which may be useful 
for the evaluation of the active learning processes.
The author of this paper attended special presentation in Barcelona, 
where results of the WCCI 2010 AL Challenge were discussed.
\end{abstract}
{\bf \small Keywords:} 
{\small unlabeled data, semi supervised learning, random sets} 

\section{Introduction}    \label{sec:intro}

Traditional supervised learning algorithms use whatever labeled data is 
provided to construct a model.
By contrast, \textit{active learning} gives the learner a degree of control by allowing 
the selection of labeled instances to be added to the training set \cite{Settles08}.  
One of the most popular methods of AL is \textit{uncertainty sampling}, 
where the learner queries the 
instance about which it has the least certainty \cite{Culotta05}, or 
\textit{query-by-committee}, where a \lq\lq committee" of models selects the instance 
about which its members most disagree \cite{Seung92}.

Based on our experience, we can divide the AL process into three main periods:
1) initial, which may be characterised by a high 
level of volatility, because of the lack of information;
2) actual, during which significant progress may be made, and 3) validation, 
which may be implemented, for example, using random sampling.

We claim that any intermediate results during an initial period are not important, because 
these results are based on insufficient evidence. 
However, the outcome of the initial period is significant as it represents a 
starting point for the 
following most important period of actual AL.

During the second period of actual AL, it is essential for the learning system to 
demonstrate speed and consistency of the improvement, and we can use a variety of methods to 
check convergence of the learning process. 
After we have found that the results are satisfactory, we can collect 
at random a large amount of labels and we can evaluate the learning trajectory 
(see Figures~\ref{fig:figure1} and ~\ref{fig:figure2}). 
Rows \lq\lq AUC", \lq\lq ALC" and \lq\lq AUC-rand", \lq\lq ALC-rand" of 
Table~\ref{tb:table1} demonstrate 
similarity between the final (official) results and the corresponding results, 
which were evaluated using random sampling.

AL may be implemented through very interesting and sophisticated methods 
\cite{Cohl96, Tong01}. 
Unfortunately, it is most unlikely that these methods would be very efficient 
in this particular Challenge because of the 
problems with the evaluation criterion (see Section~\ref{sec:alc} for more details). 

Some interesting papers including report of the 
Organisers of the WCCI 2010 AL Challenge \cite{guyon2010}  
maybe downloaded for free from the web-site of the Journal of Machine Learning 
Research\footnote{http://jmlr.csail.mit.edu/proceedings/papers/v16/}.

\begin{figure}
\begin{center}
\includegraphics[scale=0.35]{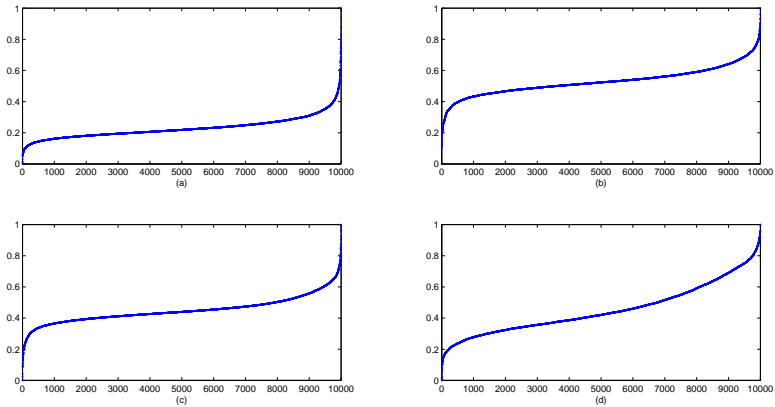}
\end{center}
\caption{\small Sorted decision functions corresponding to the set D: (a) step 2; 
(b) step 5; (c) step 12 and (d) step 16.}
\label{fig:figure3}
\end{figure}

\section{Task Description and Methods}

In many applications, including handwriting recognition, chemo-informatics, 
and text processing, large amounts of 
unlabeled data are available at low cost, but labeling examples 
(using a human expert to find the 
corresponding labels) is tedious and expensive. Hence there is a benefit 
either to use unlabeled data to improve 
the model in a semi supervised learning algorithm \cite{Kemp04, Laff08}, 
or to sample efficiently and to use human expertise for 
labeling only the most informative examples. 

Six tasks named A, B, C, D, E, and F of pool-based active learning were 
considered during the AL Challenge, 
in which large unlabeled datasets 
are available from the outset of the Challenge and the participants can 
place queries to acquire data for some 
amount of virtual cash. The participants need to return prediction values 
for all the labels every time 
they want to purchase new labels. This allows the Organisers to draw learning 
curves of prediction 
performance versus the  
amount of virtual cash spent. The participants are judged according to the 
area under the learning curves (\ref{sec:alc}). 

\begin{table}
\caption{\small Some statistical characteristics related to the Sets A-F and to the methods, 
which we used during the AL Challenge.}  \scriptsize
{\begin{tabular}{lllllll}  
\bfseries DataSet       &       \bfseries A     & \bfseries B & \bfseries C     & \bfseries D   &       \bfseries E     & \bfseries F \\
\hline
\noalign{\smallskip}
\bfseries Sample size   &       17535   &       25000   &       25720   &       10000   &       32252   &       67628   \\
\bfseries No of submissions     &       13      &       17      &       16      &       17      &       13      &       13      \\
\bfseries Used samples  &       811     &       1101    &       2301    &       531     &       661     &       900     \\
\bfseries Percentage    &       4.63$\%$        &       4.40$\%$        &       8.95$\%$        &       5.31$\%$        &       2.05$\%$        &       1.33$\%$        \\
\bfseries Last weight   &       4.32    &       4.51    &       3.48    &       4.24    &       5.61    &       6.23    \\
\bfseries Validation    &       900     &       1200    &       2000    &       1000    &       800     &       1999    \\
\hline
\noalign{\smallskip}
\bfseries Percent (positives)   &       30.95$\%$       &       17.98$\%$       &       16.71$\%$       &       39.36$\%$       &       18.76$\%$       &       27.86$\%$       \\
\bfseries $AUC_1$       &       0.5177  &       0.6304  &       0.4988  &       0.5098  &       0.5613  &       0.6179  \\
\bfseries $AUC$ &       0.8847  &       0.7323  &       0.7766  &       0.9404  &       0.7457  &       0.9853  \\
\bfseries $ALC$ &       0.4775  &       0.2834  &       0.2378  &       0.602   &       0.3689  &       0.6517  \\
\bfseries $ALC_2$       &       0.5178  &       0.3941  &       0.3415  &       0.5874  &       0.3761  &       0.7456  \\
\hline
\noalign{\smallskip}
\bfseries Percent (positives) -rand     &       20.89$\%$       &       8$\%$   &       8.85$\%$        &       25.7$\%$        &       11.75$\%$       &       7.25$\%$        \\
\bfseries $AUC$ - rand  &       0.8929  &       0.7326  &       0.7854  &       0.9334  &       0.7501  &       0.9843  \\
\bfseries $ALC$ - rand  &       0.4682  &       0.2915  &       0.2367  &       0.6038  &       0.3417  &       0.6398  \\
\end{tabular}}  \label{tb:table1}
\end{table}

\subsection{Uncertainty Sampling}

At the beginning of the AL 
Challenge\footnote{http://www.causality.inf.ethz.ch/activelearning.php}, 
we were given only one labeled (positive) sample. 
We decided to use an assumption that the data are highly imbalanced with smaller number of 
positive instances. Accordingly, we considered for the first step 50-100 random sets with 
given positive sample and other samples (assumed to be negative), 
which were selected randomly, 
but under condition that they are sufficiently distant from the given positive instance.  
The decision function was calculated as a sample average. 
We continued to use similar random sampling in 
the further steps, but the proportion of the randomly selected instances was smaller. 
After 4-6 steps we stopped using random sampling. The query for a new label was carried out  
according to the structure of the decision function, 
which was sorted in increasing order; see Figure~\ref{fig:figure3}. 

We decided that the most appropriate selection of the samples in order to make a 
query is the range, where the 
decline of the decision function is changing from the rapid to smoothed.
As a consequence, the fractions of the positive samples in our training sets were 
significantly higher compared to the fractions of the positive samples in the 
validation sets, which were collected randomly; see rows \lq\lq Percent (positive)" 
and \lq\lq Percent (positive) - rand", Table~\ref{tb:table1}.

During the initial few steps, we used as a classifier the \textit{kridge} 
function from the CLOP package. 
After collecting more labels, we started experiments with other functions: \textit{neural}  
(also from CLOP package); $GLM$, $ADA$, and $GBM$ from $R$. Also, we 
conducted feature selection 
using the Wilcoxon criterion for the Sets A, B, C and E; and a special 
likelihood-based criterion 
(binary) was used in application to the Set D. Below the level of 200-300 of the labeled 
samples, we used LOO for the evaluation and optimisation of the used parameters. 
Then, we conducted experiments using CV with 10-20 folds. The final decision function 
was computed as an ensemble of the base decision functions, where particular weights 
were defined according to the results of the evaluations with CV.

On the final phase of our experiment, we collected at random sufficiently large sample 
for the evaluation of our learning curve (the sizes of the validation samples are given 
in the row \lq\lq Validation", Table~\ref{tb:table1}).

\subsection{An ensemble constructor (a general approach)}  \label{ensemble}

\begin{definition}
An ensemble is defined as heterogeneous if the base models in an 
ensemble are generated by methodologically different learning algorithms 
(we shall consider such an ensemble in this Section). 
On the other hand, an ensemble is defined as homogeneous if the base models 
are of the same type (for example, boosting or random forest).
\end{definition}

Suppose, we have two high quality solutions, which are very different. 
Obviously, a direct sample average will not be efficient in this particular case. 

The following very simple Matlab code may be very useful in order to adjust one 
solution to the scale of the other solution without any loss of quality.

\begin{quote}  \small
\begin{verbatim}
[a1,f1]=sort(x1);   - solution N1
[a2,f2]=sort(x2);   - solution N2
[b2,g2]=sort(f2);   
x3=x1;  
for i=1:size(x1,1),   
    ii=g2(i);  
    x3(i)=a1(ii);   - adjusted solution
end;    
\end{verbatim}
\end{quote}

After the above adjustment, we can compute an ensemble solution as a linear combination:
\begin{equation} \label{eq:ens1}
   x_{ens} = \tau \cdot x_1 + (1-\tau) \cdot x_3 
\end{equation}
of the input solutions $x_1$ and $x_3,$ where $0 < \tau < 1$ is a positive weight coefficient. 
Clearly, the stronger performance of the solution $x_1$ compared to $x_3$, the bigger will be 
the value of the coefficient $\tau.$

\begin{figure}[tbh]  
\begin{center}
\includegraphics[scale=0.35]{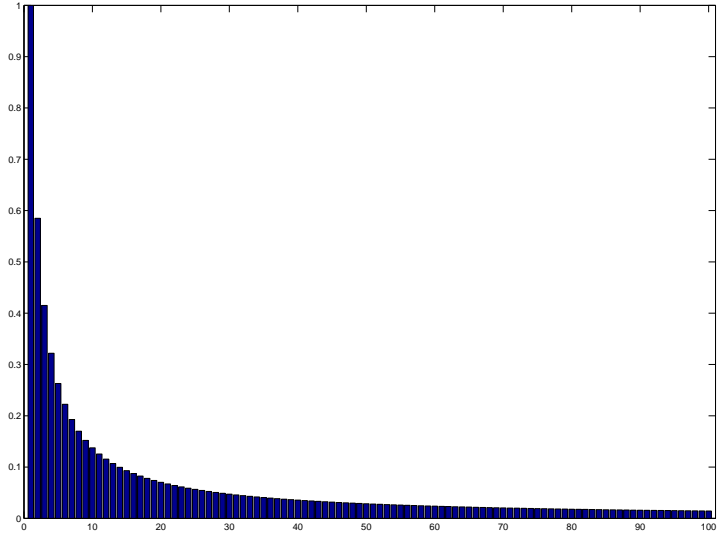}
\end{center}
\caption{\small Behavior of the weight function $w,$ which is defined in (\ref{eq:alc1}).}
\label{fig:figure4}
\end{figure}

\section{ALC Criterion}   \label{sec:alc}

Suppose, we have a sequence: $n_i, i=1,\ldots,N$ - to be the sizes of the requests, 
where $n_1 = 1$; $t_i = \sum_{j=1}^i n_j$ - the corresponding training sizes.

By definition, the area under learning curve (ALC) is
\begin{equation} \label{eq:alc}
ALC = \frac{2}{\log_2{(T)}} \sum_{i=1}^N \widehat{AUC}_i \log_2{\frac{t_{i+1}}{t_i}} - 1,
\end{equation}
where $$\widehat{AUC}_i = \frac{AUC_i +AUC_{i+1}}{2},$$ 
$\hspace{0.05in} AUC_{N+1} = AUC_N, t_{N+1} = T$ is the total available sample size 
(see row \lq\lq Sample size" in Table~\ref{tb:table1}). 

In order to simplify further notation, suppose that $n_i =n_2 +1, i=3,\ldots,N, N \geq 3.$
Then, the criterion ALC, which is given by equation (\ref{eq:alc}) 
may be rewritten in the following form, 
\begin{equation} \label{eq:alc1}
ALC \sim \widehat{AUC}_1 \log_2{(n_2 + 1)} + \sum_{i=2}^{N-1} \widehat{AUC}_i w_i + AUC_N \log_2{\frac{T}{t_N}},
\end{equation}
where 
$$w_i = \log_2{(1+\frac{1}{i-1})},$$
and we ignore in (\ref{eq:alc1}) the linear and shift coefficients as 
they do not make any difference to the classification of the learning curves.

\begin{figure}
\begin{center}
\includegraphics[scale=0.35]{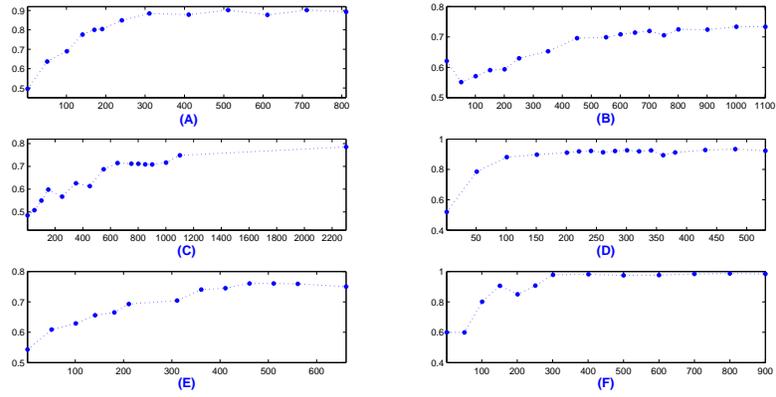}
\end{center}
\caption{\small Trajectories in terms of AUCs corresponding to the Sets A-F.}
\label{fig:figure1}
\end{figure}
\begin{figure}
\begin{center}
\includegraphics[scale=0.35]{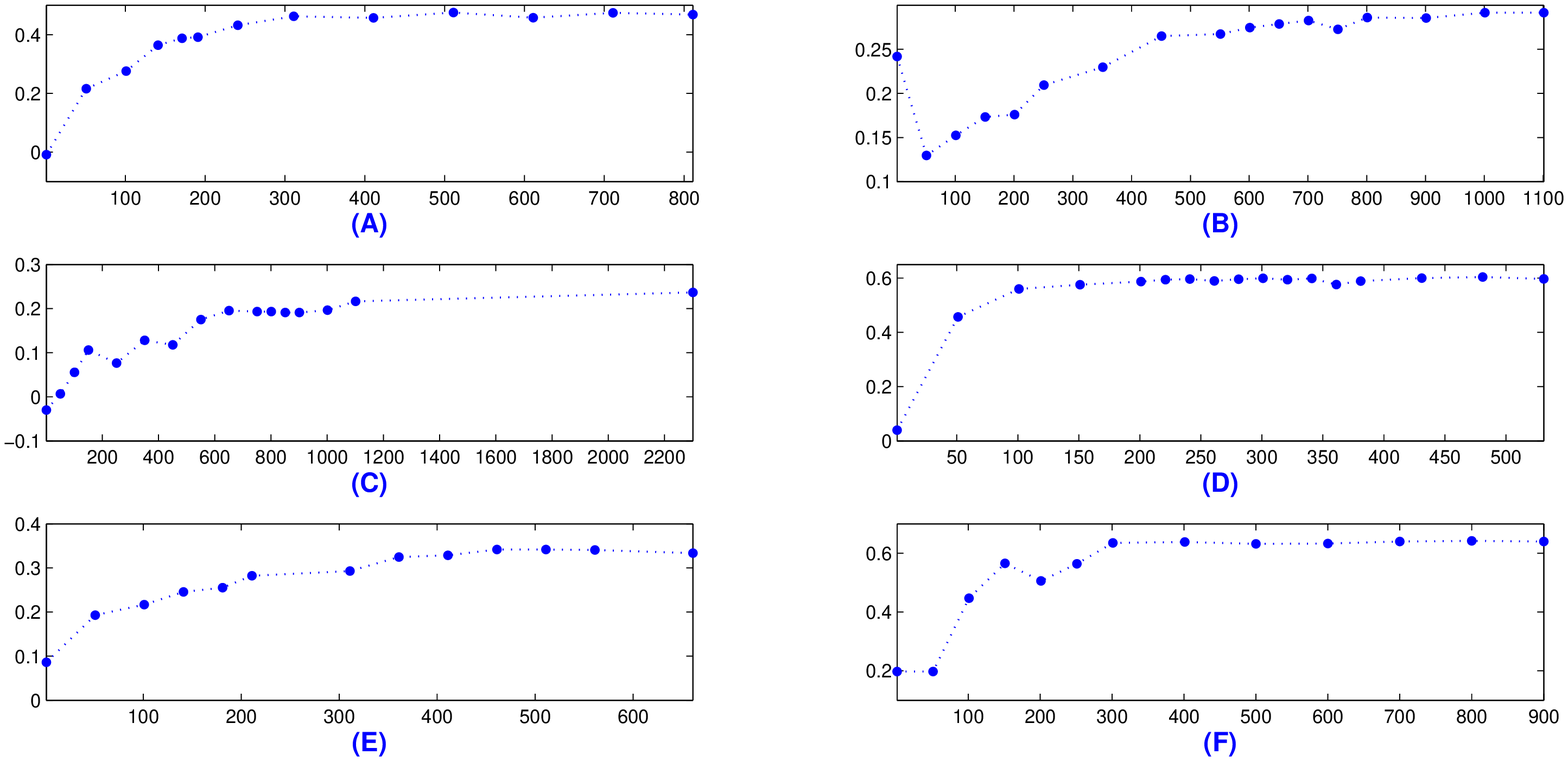}
\end{center}
\caption{Trajectories in terms of ALC's corresponding to the Sets A-F.}
\label{fig:figure2}
\end{figure}

Figure~\ref{fig:figure4} illustrates the rapid decline in the 
coefficients $w_i, i=2,\ldots,N.$
But the real winner in this particular Challenge was the first 
coefficient $w_1 = \log_2{(n_2 + 1)}.$
In all our submissions, we used $n_2 = 50,$ which corresponds to $w_1 \approx 5.67.$
Values corresponding to the last coefficient
$$w_N = \log_2{\frac{T}{t_N}}$$
are presented in the row \lq\lq Last weight", Table~\ref{tb:table1}. 
Only in the case of Set F, where we used $1.33\%$ of all available data, 
the last term is slightly more 
important compared to the first term. In all other cases the first 
term is the most important. 
Given that we did not use more that $9\%$ of all available data, 
this fact appears to be very surprising.

During the AL Competition, we acted in accordance with the natural logic 
(learn carefully with small steps). 
We had no time to  
practice with the development sets before the Competition in order to 
discover some essential features of the criterion (\ref{eq:alc}).

\begin{figure}[t]
\begin{center}
\includegraphics[scale=0.35]{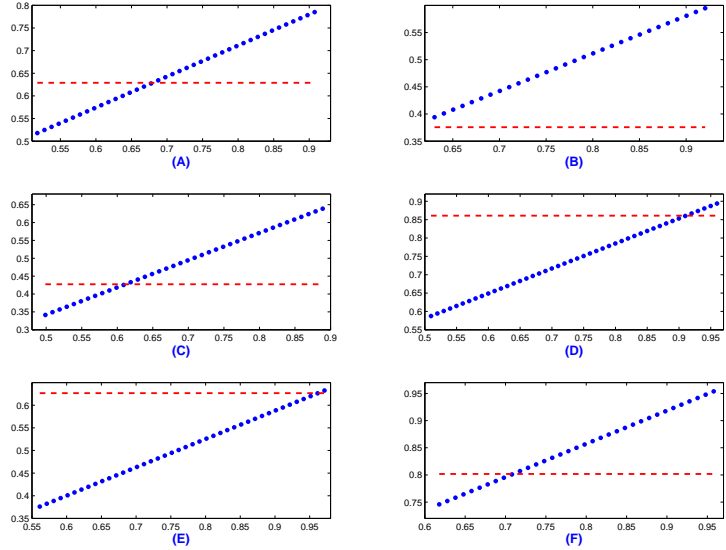}
\end{center}
\caption{\small ALCs as a function of the first entry (see for more details 
Section~\ref{sec:special}). 
The exact values of the lowest points (blue dotted lines) are given in the 
row \lq\lq $ALC_2$", Table~\ref{tb:table1}.}
\label{fig:figure5}
\end{figure}

\subsection{Illustration}

Figures~\ref{fig:figure1}(F) and ~\ref{fig:figure2}(F) illustrate the fact that the first two 
values (vertical points) of the trajectories 
corresponding to the set F are equal. This happened accidentally because we submitted 
solution from the wrong 
directory. In this particular Challenge such a mistake was very serious. 
In order to check sensitivity of the ALC evaluation criterion, we replaced the second 
value by the average of first and 
third values. As a result, the ALC grew from $0.6517$ to $0.6842.$

\subsection{Special binary case: \textit{N=2}} \label{sec:special}

After the results of the competition were released, we noticed that binary 
strategies with only 
two entries (or with the first step as a very big jump) were very popular. 
The motivation for such an approach is very simple.   
As it discussed in Section~\ref{sec:alc}, it is extremely important to maximise 
$\widehat{AUC}_1,$
because of the given evaluation criterion (\ref{eq:alc}).
 
One way to do this is through the cooperation between different teams 
(formally it is restricted, 
but cannot be controlled with full certainty).
The second way is to collect at random a very large sample, 
for example, $5-10\%$ of all available data (to ensure sufficiently large value of $AUC_2$). 
Probably, that was the best choice during this Challenge.
In fact, we have found that the criterion (\ref{eq:alc}) does not encourage, 
but discourage the process of actual AL.

Let us consider more detailed illustration for  the above fact.
In the case $N=2$ we can rewrite (\ref{eq:alc}) in the following simplified form:
\begin{equation} \label{eq:alc2}
ALC_2 = \frac{2}{\log_2{(T)}}\left( \frac{AUC_1 +AUC_2}{2}\log_2{(t_2)} + 
AUC_2 \log{\frac{T}{t_2}}\right) -1.
\end{equation}
Figure~\ref{fig:figure5} illustrates behaviour of $ALC_2$ as a function of $AUC_1,$ where the 
left (smallest) horizontal point corresponds to our initial AUC-score, values of $t_2$ 
are given in the row \lq\lq Used samples", 
values of $AUC_2$ are given in the row \lq\lq AUC", Table~\ref{tb:table1}. 

\begin{remark} We were able to observe that such dramatic simplification of the strategy 
led to improvement of the winning ALC score for Set B (we do believe that 
we will be able to reproduce our final 
result of 0.7323 using 1101 randomly selected labeled instances). 
All results corresponding to 
the binary strategy with our initial 
and final AUC scores are given in the row \lq\lq $ALC_2$", Table~\ref{tb:table1}, 
where only in the case of Set D the 
result was slightly worse compared to our final result for Set D.
\end{remark}

\begin{remark}
Normally, it is unlikely to expect that an initial AUC score based on only one 
labeled point will be better than 
0.55. We do believe, if an initial result were greater than 0.65, there must 
be some problems with the data or 
competitor who generated such result used some additional information, 
which was restricted by the rules of the Challenge. 
Our initial AUC-scores are given in the row \lq\lq $AUC_1$", Table~\ref{tb:table1}.
\end{remark}

\subsection{Optimal size of the request in the binary case} \label{sec:binary}

It will be more convenient to rewrite equation (\ref{eq:alc2}) in the following form
\begin{equation} \label{eq:alc2a}
ALC_2(t) \sim  \frac{\log_2{(t)}}{2}(AUC_1 - AUC_2(t)) + AUC_2(t) \log_2{(T)},
\end{equation}
where $ALC_2(t)$ is an increasing function of $1 \leq t \leq T$: $AUC_1=AUC_2(1).$

Suppose that based on the arguments and considerations of the previous section we decided to 
select a binary strategy, and our task is to maximise (\ref{eq:alc2a}). 

Figure~\ref{fig:figure1} illustrates that after some point the behaviour of the 
AUC graphs become nearly stable, 
and there are no sense to go further, because the above target function 
(\ref{eq:alc2a}) will decline slowly at the logarithmic rate.

The main question is how to define the optimal stopping point. 
In the case where the competitor follows correctly the second period of actual 
AL with small steps, 
this problem will be solved naturally by comparing the current and several previous solutions. 
But the competitor will face an extra penalty as a result of the small steps 
during initial period.
So in this particular Challenge the random selection of the $5\% - 10\%$ of the 
available labels 
will be fine and the most appropriate (\lq\lq the first step as a big jump").
As we can see, the competitor is strongly encouraged to avoid the most 
important period of actual AL: 
just make a big jump, and there are absolutely no any sense to do anything after that.

It is most unlikely that the Organisers would have used the evaluation criterion 
(\ref{eq:alc}) if they had had this paper in hand before.

\section{Two Proposed Evaluation Criteria}

\subsection{Formulation of the framework related to the modified criterion} \label{sec:cr1}

Our target is to produce better classification accuracy with a smaller number of 
labeled examples.

The learning process includes two subintervals (in accordance to the 
number of the used labels): 

1) The first subinterval in which the number of labeled instances is less than $\delta$ 
(for example, $\delta$ may be $1\%$ of all data available and (anyway) not more than 200). 
This subinterval will not be counted for the Challenge.

2) The second subinterval of the actual AL after $\delta.$ 
This subinterval will be counted for the Challenge.

\begin{figure}[t]
\begin{center}
\includegraphics[scale=0.35]{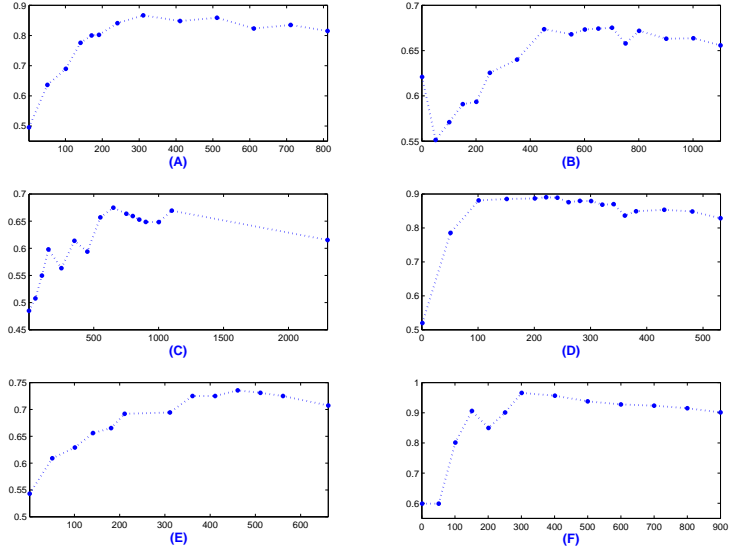}
\end{center}
\caption{\small The same trajectories as in Figure~\ref{fig:figure1}, 
evaluated with criterion (\ref{eq:crit2}).}
\label{fig:figure6}
\end{figure} 

\subsection{Motivation}

In an industrial sense, any decision making based on a small number of labeled instances 
cannot be regarded as a serious.
Any results, which are based on a small number of labeled instances, 
have no sufficient grounds to be implemented (and, probably, may only give some 
directions for further studies).

\textbf{The main idea:} it is not important how the participant will grow to the 
level of $\delta$, the actual learning and savings will be started after that.

\subsection{Second proposed criterion} 

Let us consider 
\begin{equation} \label{eq:crit2}
Q = \max_{i=1,\ldots,N} \frac{\delta \cdot AUC_i}{\delta + \alpha 
\cdot \max{( 0, t_i - \delta )}},
\end{equation}
where $\delta > 0 $ and $\alpha > 0$ are regulation parameters.

\subsection*{Example}

Similarly as in Section~\ref{sec:cr1}, 
we can select the threshold $\delta$ in (\ref{eq:crit2}) as $1\%$ of $T$ : 
$\delta = 0.01 \cdot T.$ 

Suppose that $A$ and $B$ are the expected values of AUC corresponding to 
$\delta$ and $\Delta,$ 
where $\Delta$ is $20\%$ of $T$ : $\Delta = 0.2 \cdot T$ (for example, B = 1.5 A).

The following equation follows from (\ref{eq:crit2}), 
$$\frac{A}{\delta} = \frac{B}{\delta+\alpha(\Delta - \delta)}.$$

Therefore, 
$$\alpha = \frac{\delta}{\Delta-\delta} \left( \frac{B}{A} - 1\right) = \frac{1}{38}.$$

Note that all graphs in Figure~\ref{fig:figure6} were computed using the 
same regulation parameters as in this section.

\begin{remark} The criterion (\ref{eq:crit2}) will impose equal penalties on any solution 
made within the level of $\delta.$ After that level, the penalty will grow. 
However, the quality of the solution will grow as well, and the task is not 
to stop earlier, because 
using \lq\lq big jumps" the competitors will face the risk to miss 
an optimal point. Therefore, criterion (\ref{eq:crit2}) will encourage actual AL 
with many steps, which must be reasonably small.
\end{remark}

\section{Concluding Remarks}

As it was noticed in \cite{anne2010}, if the improvement of a quantitative 
criterion such as the error rate is the main contribution of a paper, 
the superiority of a new algorithms should always be demonstrated on 
independent validation data. 
In this sense, an importance of the data mining contests is unquestionable.
The rapid popularity growth of the data mining challenges demonstrates with 
confidence that it is the best known way to evaluate different models and systems.

While the idea of AL appears to be very promising, 
we think that it is not a quite suitable subject for data mining competitions, 
because of the difficulties to check the independence of the learning processes.
As it was discussed in this paper (and during intensive email correspondence 
shortly after results were released), 
we do believe that the evaluation criterion that was used during the AL Challenge 
severely overestimated importance of the initial learning period.
The first step (based on one labeled sample) was the most important.
In the case where this step was unsuccessful, it was not possible to 
compensate the loss by the further steps. On the other hand, if we know that the 
first submission was strong, we can request a large (at random) amount of 
labels and the success at the second step (and the final success) will be guaranteed.

Our criticism is a constructive, and we have proposed two ways how to improve 
the evaluation criterion, 
which is the topic of the primary importance for any competition.

\subsection*{Acknowledgments}  

This work was supported by a grant from the Australian Research Council. 
Also, we are grateful to the Organisers of the WCCI 2010 AL data mining Contest 
for this stimulating opportunity.

\end{document}